\documentclass[letterpaper, 10 pt, conference]{ieeeconf}  

\IEEEoverridecommandlockouts                              
\usepackage{adjustbox}
\usepackage[font=small,skip=3pt]{caption}

\usepackage{epsfig} 
\usepackage{graphicx}
\usepackage{amsmath} 
\usepackage{amssymb}  
\usepackage{algorithm}
\usepackage{algpseudocode}
\usepackage{ctable}
\usepackage{multirow}
\usepackage{epstopdf}
\usepackage{color, soul}
\usepackage{colortbl}
\usepackage{tabularx}
\usepackage{pifont}

\usepackage{caption}
\usepackage{subfig}
\usepackage{adjustbox}
\usepackage{ctable}
\usepackage{multirow}
\usepackage{epstopdf}
\usepackage{color, soul}
\usepackage{colortbl}
\usepackage{tabularx}
\usepackage{booktabs}
\usepackage{pifont}
\usepackage{verbatim}

\newcommand{\vect}[1]{\boldsymbol{#1}}

\definecolor{tabrow}{rgb}{0.88,0.88,0.88}
\definecolor{Gray}{gray}{0.8}

\usepackage[utf8]{inputenc} 
\usepackage[T1]{fontenc}    
\usepackage[hidelinks]{hyperref} 
\usepackage{url}            
\usepackage{booktabs}       
\usepackage{amsfonts}       
\usepackage{nicefrac}       
\usepackage{microtype}      

\title{\LARGE \bf
Personalized Gaussian Processes for Forecasting of Alzheimer's Disease Assessment Scale-Cognition Sub-Scale (ADAS-Cog13)}

\author{
  Yuria Utsumi$^1$, Ognjen (Oggi) Rudovic$^1$, Kelly Peterson$^1$,  Ricardo Guerrero$^2$ and Rosalind W. Picard$^1$\\
  $^1$Massachusetts Institute of Technology\\
  $^2$Imperial College London\\
  {\tt\small \{yutsumi,orudovic,kellypet,roz\}@mit.edu, reg09@imperial.ac.uk}
}

\begin{document}

\maketitle
\thispagestyle{empty}
\pagestyle{empty}

\begin{abstract}

In this paper, we introduce the use of a personalized Gaussian Process (pGP) model to predict per-patient changes in ADAS-Cog13 -- a significant predictor of Alzheimer's Disease (AD) in the cognitive domain -- using data from each patient's previous visits, and testing on future (held-out) data. We start by learning a population-level model using multi-modal data from previously seen patients using a base Gaussian Process (GP) regression. The pGP is then formed by adapting the base GP sequentially over time to a new (target) patient using domain adaptive GPs \cite{eleftheriadis2017}. We extend this personalized approach to predict the values of ADAS-Cog13 over the future 6, 12, 18, and 24 months. We compare this approach to a GP model trained only on past data of the target patients (tGP), as well as to a new approach that combines pGP with tGP.  We find that this new approach (pGP+tGP) leads to significant improvements in accurately forecasting future ADAS-Cog13 scores.

\end{abstract}

\section{Introduction}
Recently, the view on Alzheimer’s Disease (AD) diagnosis has shifted towards a more dynamic process in which clinical and pathological markers evolve gradually before diagnostic criteria are met. Given the wide variability in data available per subject, inherent per-person differences, and the slowly changing nature of the disease, accurate prediction of AD progression is a significant, difficult challenge. The Alzheimer's Disease Assessment Scale-cognition sub-scale (ADAS-Cog)~\cite{mohs1997development} is the most widely used general cognitive measure in clinical trials of AD~\cite{skinner2012alzheimer}. While it was developed as an outcome measure for dementia interventions, its primary purpose was to be an index of global cognition in response to antidementia therapies. The ADAS-Cog assesses multiple cognitive domains including memory, language, praxis, and orientation~\cite{skinner2012alzheimer}. Because ADAS-Cog has proven important for target clinical assessments, in this paper we focus on a machine learning method that can successfully forecast the future values of this score for previously seen subjects. Specifically, we use the modified ADAS-Cog 13-item scale~\cite{mohs1997development}, which includes all
original ADAS-Cog items, with the addition of a number cancellation task and a delayed free recall task, 
for a total of 85 points. As in the parent instrument, higher scores indicate greater severity. 
The purpose of these additional items was to increase the number of cognitive domains and range of symptom severity without a substantial increase in the time required for administration. 

\begin{figure}[t]
\centering
\caption{{\small {\bf Personalized GPs.} The population model is first trained using all past visits data of $N$ subjects ($x^{TR},y^{TR}$), where the time difference between two visits is 6 months. The model personalization to the target subject $(N+1)$ is then achieved by sequentially adapting the model predictions of the future ADAS-Cog13 scores $y_{t+1:t+4}$ (using the posterior distribution of GPs - $f_{GP}$), informed by the visits data up to time step $t$. The shaded fields in the output represent the time points for which no visit data is available for a given subject.}}
\label{overview}
\includegraphics[scale=.275]{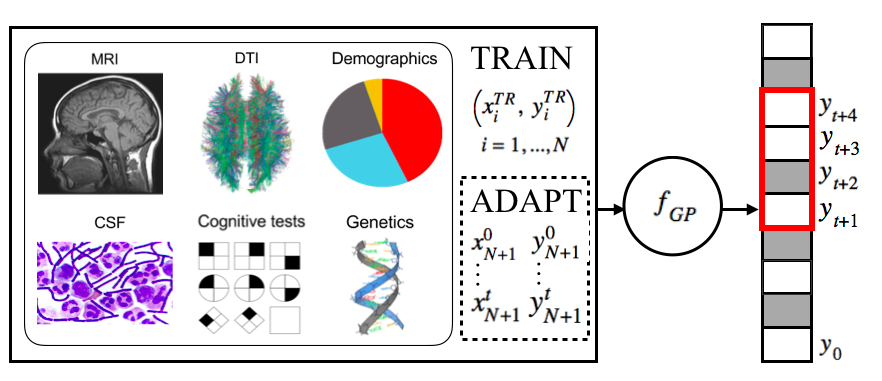}
\end{figure}
One of the main challenges in clinical assessment of subjects at risk of developing AD is the ability to accurately predict the future cognitive scores of interest for clinical trials, for instance, which aim to find the most effective treatments for AD. For this, an automated approach that could forecast the future cognitive scores such as ADAS-Cog13 would bring a great value for the assessment procedure, and has potential to significantly improve the efficacy and efficiency of clinical trials, which are typically lengthy and expensive. For example, out of hundreds of clinical trials, costing billions of dollars, fewer than $1\%$ have proceeded to the regulatory approval stage and none have managed to prove a disease-modifying effect \cite{tadpole2017,cummings2006}. More successes depend on an improved ability to accurately identify subjects at early ages of the disease where treatments are most likely to be effective. Thus, developing models with improved ability to automatically predict subjects' future AD-related metrics indicating disease progression -- and to do so as early as possible, especially before the emergence of clinical symptoms -- is an important step towards this end. Furthermore, accurate prediction of symptom onset in the time window of 6 to 24 months is critical to participant selection and formation of clinical trials. Thus, having access to accurate future estimates of the progression of cognitive scores such as ADAS-Cog13 within this time-frame is of great importance. 

In this paper, we use data from the Alzheimer’s Disease Neuroimaging Initiative (ADNI)\cite{weiner2017}, and, specifically, the dataset processed for the TADPOLE Challenge~\cite{tadpole2017}, to address the problem of forecasting ADAS-Cog13. These data are highly heterogeneous and multi-modal, and include imaging (MRI, PET), cognitive scores, CSF biomarkers, genetics, and demographics (e.g. age, gender, race) \cite{tadpole2017}. Although the heterogeneous nature of this dataset lends itself to building powerful, informative multi-modal models, the dataset itself is very sparse, with different combinations of features missing for different subjects. Partial records are also missing for roughly $80\%$ of subjects \cite{campos2015}. Moreover, given the wide variability in available data per subject, inherent per-person differences, and the slowly changing nature of the disease, accurate forecasting of cognitive decline and related measures of disease progression is a significant and difficult challenge.   

To tackle the challenges mentioned above, we focus on machine learning models that can easily adapt to each individual and perform forecasting of his/her cognitive score ADAS-Cog13. More specifically, we investigate the effects of the model personalization for the forecasting task using the framework of Gaussian Processes (GP)~\cite{rasmussen2006gaussian}. This non-parametric probabilistic framework offers great modeling flexibility by providing not only the predictions of future cognitive scores but also uncertainty in these predictions, which enables better quantification of the forecasting performance. For the forecasting task, we first
employ the personalized Gaussian Process model (pGP)~\cite{peterson2017personalized} to simultaneously predict ADAS-Cog13 scores four steps ahead in time, given previous assessment data of target subjects (see Fig.\ref{overview}). We forecast using a period of four time steps ahead (6, 12, 18, and 24 months) due to its relevance for clinical trials, as well as because our analysis of the auto-correlations in ADAS-Cog13 scores (based on TADPOLE data) revealed their strong correlation ($>80\%$) within the used forecasting window. 

We evaluate our approach on a cohort of 100 subjects from the ADNI dataset to predict future ADAS-Cog13 scores of target subjects using data from each subject’s previous visits. We define a subject visit as data collected at a single time point during ADNI. We start by learning a population-level model using multi-modal data of previously-seen (source) subjects using the base GP regression (sGP). Then, this model is adapted sequentially over time to a new (target) subject using the notion of domain adaptive GPs~\cite{liu2015bayesian,eleftheriadis2017}. We extend this personalization approach, which we denote as pGP, for the forecasting task by learning to predict ADAS-Cog13 up to four steps ahead (24 months), and show that it outperforms sGP on a large majority of target subjects. We also compare this model with the GP model trained only on the data of target subjects (tGP). Surprisingly, this model, trained with a much smaller set of data, outperforms the sGP model and also pGP on a number of subjects. To leverage this, we combine the pGP and tGP models, and find that the combination of the two achieves the best performance on the forecasting task. 

\section{Related Work}
Here, we briefly review related work on forecasting of cognitive scores and clinical status for AD assessment, with the focus on the ADNI dataset. Most existing approaches (e.g.,~\cite{schmidt-richberg16,guerrero2016,gavidia2017}) focus on modeling subjects based on their clinical status (CS): cognitively normal (CN), mild cognitive impairment (MCI) and Alzheimer's Disease (AD). However, the majority of these focus on modeling biomarkers at the population level; for instance, estimating typical trajectories of markers over the full course of the disease to estimate current disease progress and progression rate \cite{schmidt-richberg16,guerrero2016}. Guerrero et al.~\cite{guerrero2016} used mixed effects modeling to derive global and individual biomarker trajectories for a training population, which was later used to instantiate subject-specific models for unseen subjects. Some of the modeling techniques \cite{guerrero2016,schmidt_richberg15_a,schmidt-richberg16} require cohorts with known disease onset and are prone to bias due to the uncertainty of the conversion time.

When it comes to forecasting of clinical status, several authors attempted predicting target scores for longer time windows. For example, several recent works have explored the use of multiple, multi-modal predictors in combination with various machine learning techniques (e.g. SVMs, neural nets, GPs, etc.) to predict conversion from MCI to AD for various future time periods, e.g. from 1-5 years after baseline assessment~\cite{long2017,minhas2017}. Likewise, the BrainAGE framework has been proposed for predicting MCI-AD conversion within 3 years of follow-up \cite{gaser2013}. In addition, to predict conversion within time windows of up to 2 years (short-term converter) and 2-4 years (long-term converter), \cite{pereira2017} proposed a stepwise learning approach, where the learned model first predicts whether a subject converts to dementia, or remains stable, and then predicts the more likely progression window (short-term or long-term conversion). More recently, \cite{grassi2018a} evaluated different machine learning models in order to develop an algorithm for a 3-year prediction of conversion to AD in MCI and Pre-MCI subjects based only on non-invasive and effectively collectible predictors.

However, most of these works attempted forecasting of changes in subjects' CS, which deals with a limited number of future outcomes (i.e., either binary or a three-class). By contrast, we aim to forecast ADAS-Cog13, defined on a more fine-grained scale (85 levels), which poses a more challenging machine learning problem. While in our recent work~\cite{peterson2017personalized}, we investigated forecasting of ADAS-Cog13 along with other scores (CDSRB, CS, and MMSE), we did so for one step ahead (6 months). In this work, we attempt forecasting of up to 24 months ahead by focusing on ADAS-Cog13.

\section{Forecasting with Personalized GPs}
{\bf Notation.} We consider a supervised setting, where $\vect{X}^{(s)} = \{\vect{x}^{(s)}_{n_s}\}_{n_s=1}^{N_s}$ represents the multi-modal input features of $N_s$ subjects from the ADNI dataset used to train the population model. The outcome score for each individual subject visit is stored as $\vect{y}^{(s)}_{n_s}=\ 
\text{ADAS-Cog13}^{(s)}_{n_s}\in\text{(0-85)}$, and $\vect{Y}^{(s)}=\{\vect{y}^{(s)}_{n_s}\}_{n_s=1}^{N_s}$. Furthermore, each subject is represented by data pairs: $\{ \vect{x}^{(s)}_{n_s},\vect{y}^{(s)}_{n_s}\}$, where $\vect{x}^{(s)}_{n_s}=\{x_1,..,x_{t}\}$ contains the input features up to visit $t$, and $\vect{y}^{(s)}_{n_s}=\{y_{w}\}$ the corresponding scores for the future visits $w$=$t$+$1$:$t$+$4$. The total number of visits is denoted by $T$. For notational convenience, we drop dependence on $n_s$, and use $\{x_{t}^{(s)},y_{w}^{(s)}\}_{t=1:T-1}$ as data pairs for training the (population) forecasting model\footnote{From $t=T-3:T$, we forecast only the remaining scores.}. Since some subjects missed certain visits and not all biomarkers were recorded at every visit, we fill in their missing values using their nearest available past visit; however, no data of future visits is used in training.
 
 {\bf Population-level GP.} We first build the population-level forecasting model using data of training (source) subjects $\{\vect{X}^{(s)},\vect{Y}^{(s)}\}$. The goal of the forecasting model is to predict a subject's future ADAS-Cog13 score as output $y_{w}^{(s)}$ from data of previous visits, $x_{t}^{(s)}$ and $y_{t}^{(s)}$. To this end, we use the auto-regressive GP~\cite{candela2003propagation}, which learns the following forecasting function:
\begin{equation}
 y_{w}^{(s)} = f^{(s)}(x_{t}^{(s)};y_{t}^{(s)}) + \epsilon_{t}^{(s)},
\end{equation}
where $\epsilon_t^{(s)}\sim \mathcal{N}(0, \sigma^2_{s})$ is i.i.d. additive Gaussian noise. Following the framework of GPs~\cite{rasmussen2006gaussian}, we place a prior on the functions $f^{(s)}$. This gives rise to the joint prior $p(\vect{Y}^{(s)}_{2:T}|\{\vect{X}^{(s)},\vect{Y}^{(s)}\}_{1:T-1}) = \mathcal{N}(\vect{Y}^{(s)}_{2:T}|\vect{0}, \vect{K}^{(s)}_{1:T-1})$, where the elements of $\vect{K}^{(s)}_{1:T-1} = k^{(s)}(\{\vect{X}^{(s)},\vect{Y}^{(s)}\}_{1:T-1},\{\vect{X}^{(s)},\vect{Y}^{(s)}\}_{1:T-1})$ are computed using radial basis function (RBF) -isotropic kernel. 
The kernel parameters $\vect{\theta}$ were chosen to minimize the negative log-marginal likelihood: $-\log p(\vect{Y}^{(s)}_{2:T}|\{\vect{X}^{(s)},\vect{Y}^{(s)}\}_{1:T-1}, \vect{\theta})$. Given data from the visit at time $t$ of a {\it new} subject, $\vect{x}_\ast^{(s)}=\{x_{t}^{(s)},y_{t}^{(s)}\}$, the population GP predictive distribution provides the mean and variance forecasts of the cognitive scores $y_{w}^{(s)}$ as:
\begin{align}
\small
\label{post_mu}\vect{\mu}_{w}^{(s)} &= {\vect{k}_\ast^{(s)}}^T (\vect{K}^{(s)} + \sigma_s^2\vect{I})^{-1}\vect{Y}^{(s)} \\ 
\label{post_s}\vect{V}_{w}^{(s)} &= k_{\ast\ast}^{(s)} - 
{\vect{k}_\ast^{(s)}}^T (\vect{K}^{(s)} + \sigma_s^2\vect{I})^{-1}
\vect{k}_\ast^{(s)},
\end{align}
where {\small $\vect{k}_\ast^{(s)} = k^{(s)}(\vect{X}^{(s)}, \vect{x}_\ast^{(s)})$} and {\small $k_{\ast\ast}^{(s)} = k^{(s)}(\vect{x}_\ast^{(s)}, \vect{x}_\ast^{(s)})$}, and the distribution mean is used as a point estimate of target outputs. We denote $\vect{\mu}_{w}^{(s)}=\vect{\mu}_{w}^{(s)}(\vect{x}_\ast^{(s)})$ and $\vect{V}_{w}^{(s)}=\vect{V}_{w}^{(s)}(\vect{x}_\ast^{(s)})$, and refer to this setting as sGP (source GP).

{\bf Personalized GP (pGP).}
\label{pgp}
We extend the approach of domain adaptive GPs (DA-GP)~\cite{liu2015bayesian,eleftheriadis2017} to personalize the population GP model to target subjects. This is achieved by sequentially adapting the GP posterior for the test subject using the data of his/her past visits, to forecast the future ADAS-Cog13 scores $y_{w}^{(p)}$. This is achieved by using the obtained posterior distribution of the source (population) data and data of the target subject up to visit $t$, to obtain a prior for the GP of the future data for the subject: $p(\vect{Y}^{(p)}_{w}|\{\vect{X}^{(p)},\vect{Y}^{(p)}\}_{1:t}, \mathcal{D}^{(s)}, \vect{\theta})$. This prior is used to correct the posterior distribution derived above to account for the previously seen data of the target subject. Formally, the conditional prior on the target subject data (given the source data) is obtained by applying Eqs.~(\ref{post_mu}--\ref{post_s}) on $\{\vect{X}^{(p)},\vect{Y}^{(p)}\}_{1:t}$ to obtain $\label{prior_mut}\vect{\mu}_{_{w}}^{(p|s)}$ and $\label{prior_st}\vect{V}_{w}^{(p|s)}$. Given the prior and a test input $\vect{x}_\ast^{(p)}=\{x_{t}^{(p)},y_{t}^{(p)}\}$, the correct form of the adapted posterior after observing the target subject data at visit $t$ is:\\
\scalebox{1}{\parbox{1\linewidth}{%
\begin{align}
\small
\label{post_muad}\vect{\mu_{w}^{(p)}} &= \vect{\mu}_{w}^{(s)} +  {\vect{V}_{w}^{(p|s)}}^T (\vect{V}_{1:t}^{(p|s)} + \sigma_s^2\vect{I})^{-1}(\vect{Y}^{(p)}_{1:t} - \vect{\mu}_{1:t}^{(p|s)}) \\ 
\label{post_sad}\vect{V_{w}^{(p)}} &= \vect{V_{w}^{(s)}} - 
{\vect{V}_{w}^{(p|s)}}^T (\vect{V}_{1:t}^{(p|s)} + \sigma_s^2\vect{I})^{-1}
\vect{V}_{w}^{(p|s)},
\end{align}
}}\\
with {\small $\vect{V}_{w}^{(p|s)} = k^{(s)}(\vect{X}^{(p)}, \vect{x}_\ast^{(p)}) - 
{k^{(s)}(\vect{X}^{(s)}, \vect{X}^{(p)})}^T (\vect{K}^{(s)} + \sigma_s^2\vect{I})^{-1} k^{(s)}(\vect{X}^{(s)}, \vect{x}_\ast^{(p)})$}. Eqs.~(\ref{post_muad}--\ref{post_sad}) show that final forecast by the pGP is the combination of the population-model forecast, plus a correction term\footnote{For $t$=$1$, the population model is used instead.}. The latter shifts the mean toward the distribution of the target subject and improves the model's confidence by reducing its predictive variance. Note that we use the shared covariance function for simultaneous forecasting of $y_{w}$. Consequently, the model assigns the same variance ($\vect{V}_{w}$) to each future forecast in $w$.

{\bf Target GP (tGP).}
We also build the GP model 
from the observed data of the target subject (the same data used for the adaption in the pGP). However, since this data set is of insufficient size to learn the GP parameters -- the GP would easily overfit -- we use the kernel parameters of the population model (sGP). Unlike the sGP, whose covariance matrix is fixed (learned from previously seen source subjects), the tGP continually updates its covariance matrix as more past data (up to $t$) of a target subject becomes available. Nevertheless, the inference procedure remains the same. Formally, this is given by:  
\scalebox{1}{\parbox{1\linewidth}{%
\begin{align}
\small
\label{post_mut}\vect{\mu_{w}^{(t)}} &= {\vect{k}_{*}^{(t)}}^T (\vect{K}_{1:t}^{(t)} + \sigma_s^2\vect{I})^{-1}\vect{Y}^{(t)}_{1:t} \\ 
\label{post_st}\vect{V_{w}^{(t)}} &= k_{**}^{(t)} - 
{\vect{k}_{*}^{(t)}}^T (\vect{K}_{1:t}^{(t)} + \sigma_s^2\vect{I})^{-1}
\vect{k}_{*}^{(t)}.
\end{align}
}}\\
For exact derivation of the model equations, see~\cite{liu2015bayesian,eleftheriadis2017}.

\section{RESULTS}

Data used in this study were collected from the ADNI database \href{http://adni.loni.usc.edu/}{(adni.loni.usc.edu)}. We downloaded the standard dataset processed for the TADPOLE Challenge~\cite{tadpole2017}; this dataset represents 1,737 unique subjects and was created from the \href{https://adni.bitbucket.io/index.html}{ADNIMERGE} spreadsheet, to which regional MRI (volumes, cortical thickness, surface area), PET (FDG, AV45, AV1451), DTI (regional means of standard indices) and cerebrospinal fluid (CSF) biomarkers were added. From this data, we use all features provided except for clinical status (CS) and normalized 

ventricle volumes (ICVn)\footnote{In the TADPOLE Challenge, AdasCog-13, CS and ICVn are treated as target outputs, so in this work we excluded the latter two, and other cognitive scores (CDRSB and MMSE) as predictors in our model.}, to construct a multi-modal feature set consisting of six modalities: demographics (6 features), genetics (3 features), cognitive tests (9 features), CSF (3 features), MRI (365 features), and DTI (229 features). Due to sparseness, we excluded PET data entirely. However, in the future, we plan to extend this work and incorporate all three predictors (ADAS-Cog13, CS, and normalized ventricular volume). In the experiments reported, we used a cohort of 100 subjects with more than $10$ visits and missing no more than $82.5\%$ of the features.

To evaluate performance, we ran a 10-fold subject-independent cross-validation. All the input features were z-normalized (zero mean, unit variance). As performance metrics, we report the mean absolute error (MAE) and weighted error score (WES), as suggested by the TADPOLE challenge \cite{tadpole2017}. MAE is the standard measure for regression models and purely measures accuracy of prediction ignoring confidence. However, we also exploit the confidence in each prediction, as this is commonly used when evaluating forecasting models. Thus, we report WES, which incorporates the confidence estimates for the prediction into the score. We compute WES using the $50\%$ confidence intervals ($CI$) obtained from the predictive distribution of the GPs as $CI=[{\mu _w} \pm 0.67{({V_w})^{1/2}}]$~\cite{tadpole2017}. Note that we used the imputed values (i.e., ADAS-Cog13 scores) for the missing visits to form the equally spaced visits (6,12,18 and 24 months), but the reported errors are computed on the existing visits only.

We report the results obtained using the following four model variations: GP models learned from source subjects only (sGPs), personalized versions of population-level GP models (pGPs), and the GP models learned using only data of target subjects (tGPs) up to time $t$. Since we found that tGP performed very well (via a validation on source subjects mimicking the forecasting task), we defined a joint GP model that combines the personalized approach with target subject data (pGP + tGP). We experimented using variance-based weighting as in distributed GP approaches~\cite{deisenroth2015distributed,eleftheriadis2017}. However, tGP tended to overestimate the variance due to the small amount of per-person data used to build this model. By averaging the pGP and tGP forecasts, we obtained the best performance. Since both pGP and tGP provide a Gaussian predictive distribution, the distribution of the average model is also Gaussian ${\mathcal N}(\frac{1}{2}(\vect{\mu _w^{(p)}} + \vect{\mu _w^{(t)}}),\frac{1}{4}(\vect{V_w^{(p)}} + \vect{V_w^{(t)}}))$. All GP models were trained using GP toolbox~\cite{rasmussen2006gaussian}.

In Fig. 2, we compare the trend of average MAE among the four models, evaluated up to the 5th, 10th, 15th, and 21st visits, across four time steps. Generally, MAE for sGP, pGP and pGP+tGP models follows a decreasing trend, with sGP having the highest error, and pGP outperforming tGP as more visits' data become available. The joint pGP+tGP model tends to mirror this error trend and mirrors the error magnitude of tGP. Essentially, the joint model inherits the lower error rate of tGP and the general trend of decreasing error with visit number of pGP, which explains its best overall performance, relative to the other models.  We also observe a crossover of pGP+tGP with pGP and tGP for $t+1$ time step. For $t+1$ time step, pGP initially performs worse relative to pGP+tGP; however, pGP outperforms the joint model after 17 visits (see Fig. 2a). This also reflects the fact that, for subjects for whom the input features are very different from source subjects (a large covariate shift), the adaptation approach (pGP) is not very effective, and using tGP results in better individual performance. However, our results demonstrate that using the combination of the two is most effective.  

Table 1 compares the error metrics for the four models across four time steps. While pGP performs better at earlier time steps, tGP dominates at later time steps. This is expected; as more data of a target subject is available, tGP becomes more consistent in its predictions on the target subject. Again, by combining pGP and tGP into a joint model, we achieve the best overall performance in terms of MAE. In addition, from the WES scores computed using the estimated variance of the forecasts, we note that both sGP and pGP produce more confident forecasts compared to tGP, which is a consequence of a small set of data being used to compute the covariance function of the tGP model.

In Fig. 3, we compare the results of the four models and report the average results for each forecasting step. Specifically, in Fig. 3a, we compare the sGP with the joint pGP+tGP model. We see that the combined pGP+tGP model improves per subject performance largely, when compared to the sGP model alone. Likewise, comparing this joint model with only the target model (pGP), as shown in Fig. 3b, we see the difference in performance is also pronounced. Finally, from Fig. 3c we see the joint model outperforms the target model (tGP) for around half of the subjects. This improvement in performance (as can also be seen from Table 1) is due to the personalization and learning of the target GPs, compared to learning the GP from source subjects only.

\begin{table}[h]
\caption{Comparison of the proposed forecasting GP models. We report mean$\pm$SD of the 10-fold person independent cross-validation of the models, and for each forecasting step. The statistically significant differences (the paired t-test with $p=0.05$) between the two best performing models at the time are marked with *.}
\label{tab_res}
\centering
\begin{center}
\begin{adjustbox}{width=0.488\textwidth}
\begin{tabular}{l|ccccc}
  \hline
    \rowcolor{Gray}
\textbf{Models}        && &\textbf{MAE}& &\\
\hline
{}              & \text{$t+1$}& \text{$t+2$} & \text{$t+3$}& \text{$t+4$} & \text{Avg.}\\
\hline
\toprule      
\textbf{sGP} &3.86$\pm$0.37 &4.39$\pm$0.52 &4.72$\pm$0.44 &5.04$\pm$0.62 &4.50$\pm$0.66\\
\textbf{pGP} &{\bf 3.71$\pm$0.30*} &4.05$\pm$0.29 &4.54$\pm$0.36 &4.40$\pm$0.56 &4.18$\pm$0.51\\
\textbf{tGP} &4.31$\pm$0.56 &4.00$\pm$0.40 &4.63$\pm$0.38 &4.17$\pm$0.49 &4.28$\pm$0.52\\
\textbf{pGP + tGP} &3.76$\pm$0.42 &{\bf 3.72$\pm$0.28*} &{\bf 4.33$\pm$0.35*} &{\bf 3.88$\pm$0.40*} &{\bf 3.92$\pm$0.44*}\\
\bottomrule  
  \hline
    \rowcolor{Gray}
&& &\textbf{WES}& &\\
\hline
\textbf{sGP} &3.63$\pm$0.42 &4.09$\pm$0.49 &4.62$\pm$0.69 &4.57$\pm$0.66 &4.23$\pm$0.70\\
\textbf{pGP} &{\bf 3.62$\pm$0.41*} &{3.99$\pm$0.42} &4.57$\pm$0.68 &4.39$\pm$0.64 &{4.14$\pm$0.66}\\
\textbf{tGP} &4.14$\pm$0.41 &3.99$\pm$0.35 &{4.77$\pm$0.62} &{4.11$\pm$0.52} &4.25$\pm$0.57\\
\textbf{pGP + tGP} &3.65$\pm$0.55 &{\bf 3.63$\pm$0.28*} &{\bf 4.38$\pm$0.58*} &{\bf 3.86$\pm$0.46*} &{\bf 3.88$\pm$0.57*}\\
\bottomrule 
\end{tabular}
\end{adjustbox}
\end{center}
\end{table}

\begin{figure}[!tbp]
  \centering
  \subfloat[]{\includegraphics[width=0.24\textwidth]{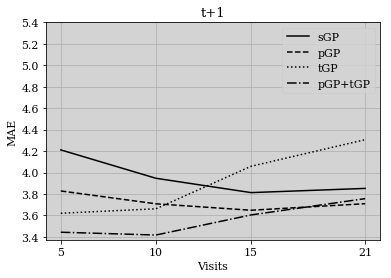}\label{fig:f1}}
  \hfill
  \subfloat[]{\includegraphics[width=0.24\textwidth]{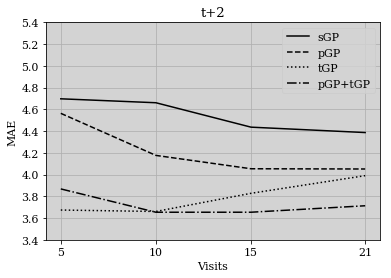}\label{fig:f2}}
   \hfill
   \subfloat[]{\includegraphics[width=0.24\textwidth]{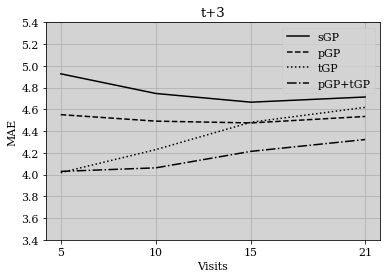}\label{fig:f3}}
  \hfill
   \subfloat[]{\includegraphics[width=0.24\textwidth]{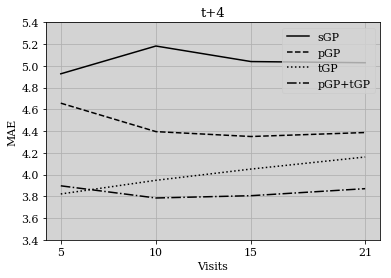}\label{fig:f4}}
  \caption{Line graphs showing trend of average MAE of each model and time step, evaluated up to 5th, 10th, 15th, and 21st visits.}
\end{figure}

\begin{figure*}[!tbp]
  \centering
  \subfloat[]{\includegraphics[width=0.33\textwidth]{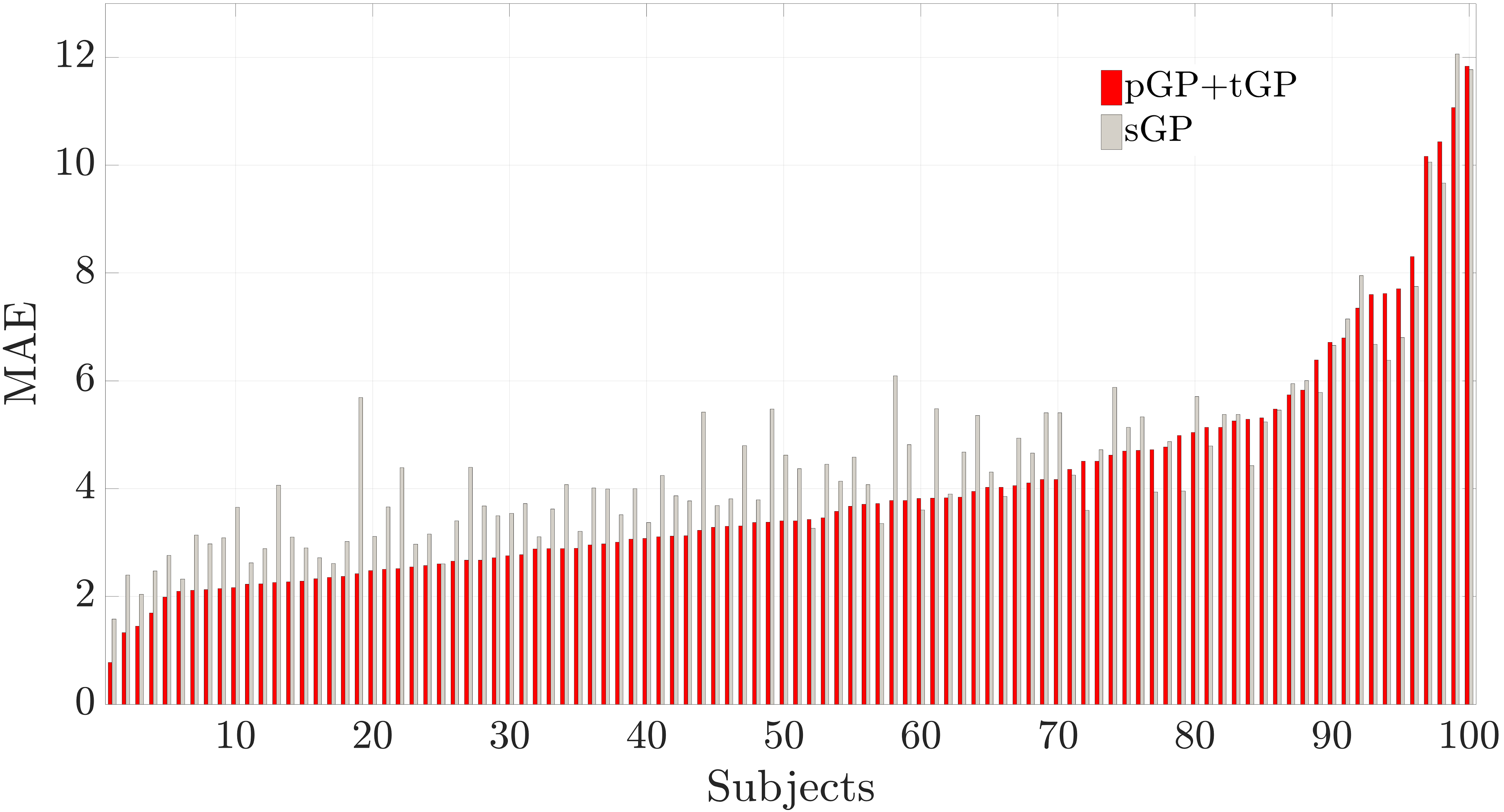}\label{fig:f4}}
  \hfill
  \subfloat[]{\includegraphics[width=0.33\textwidth]{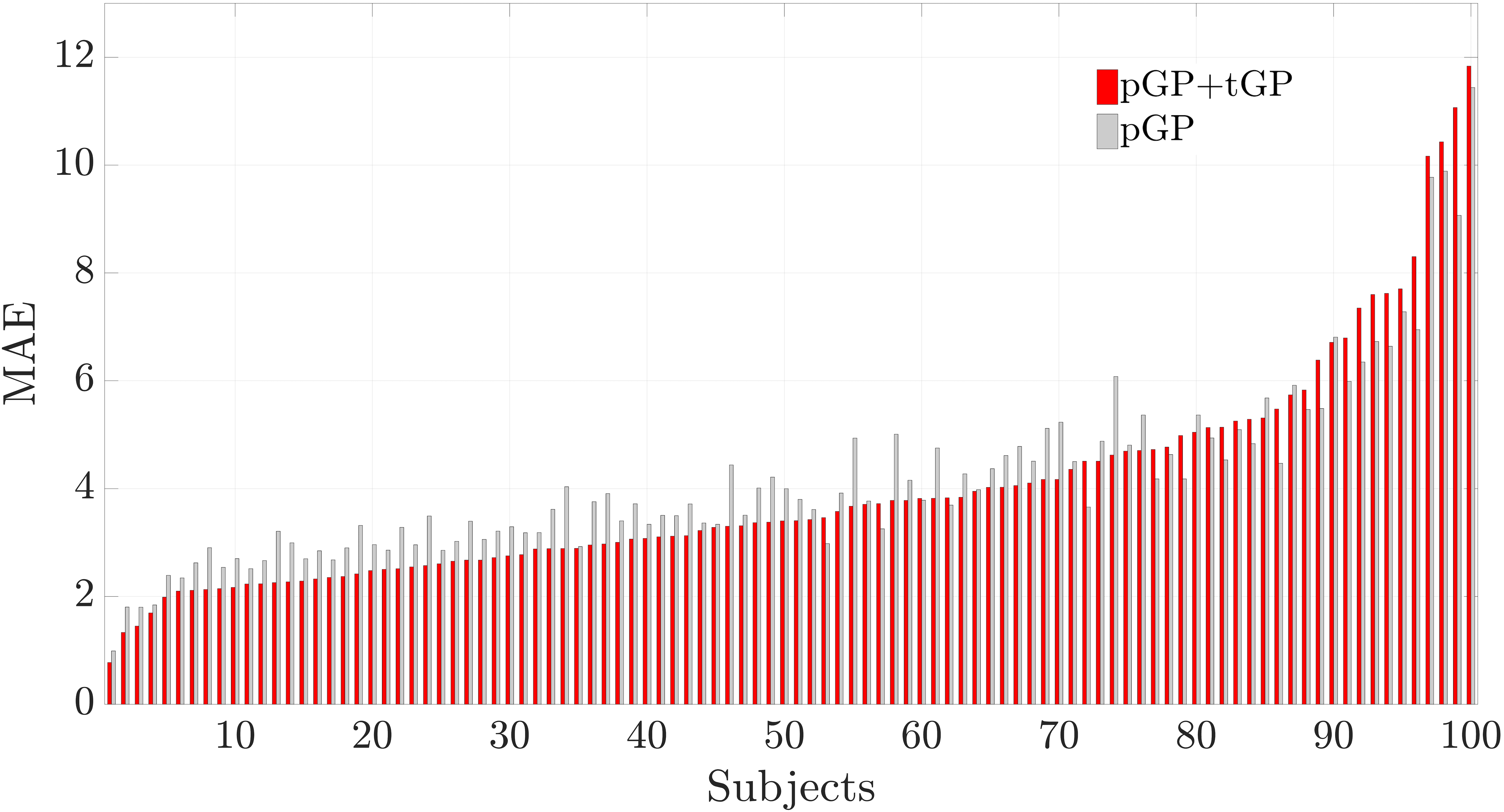}\label{fig:f5}}
   \hfill
   \subfloat[]{\includegraphics[width=0.33\textwidth]{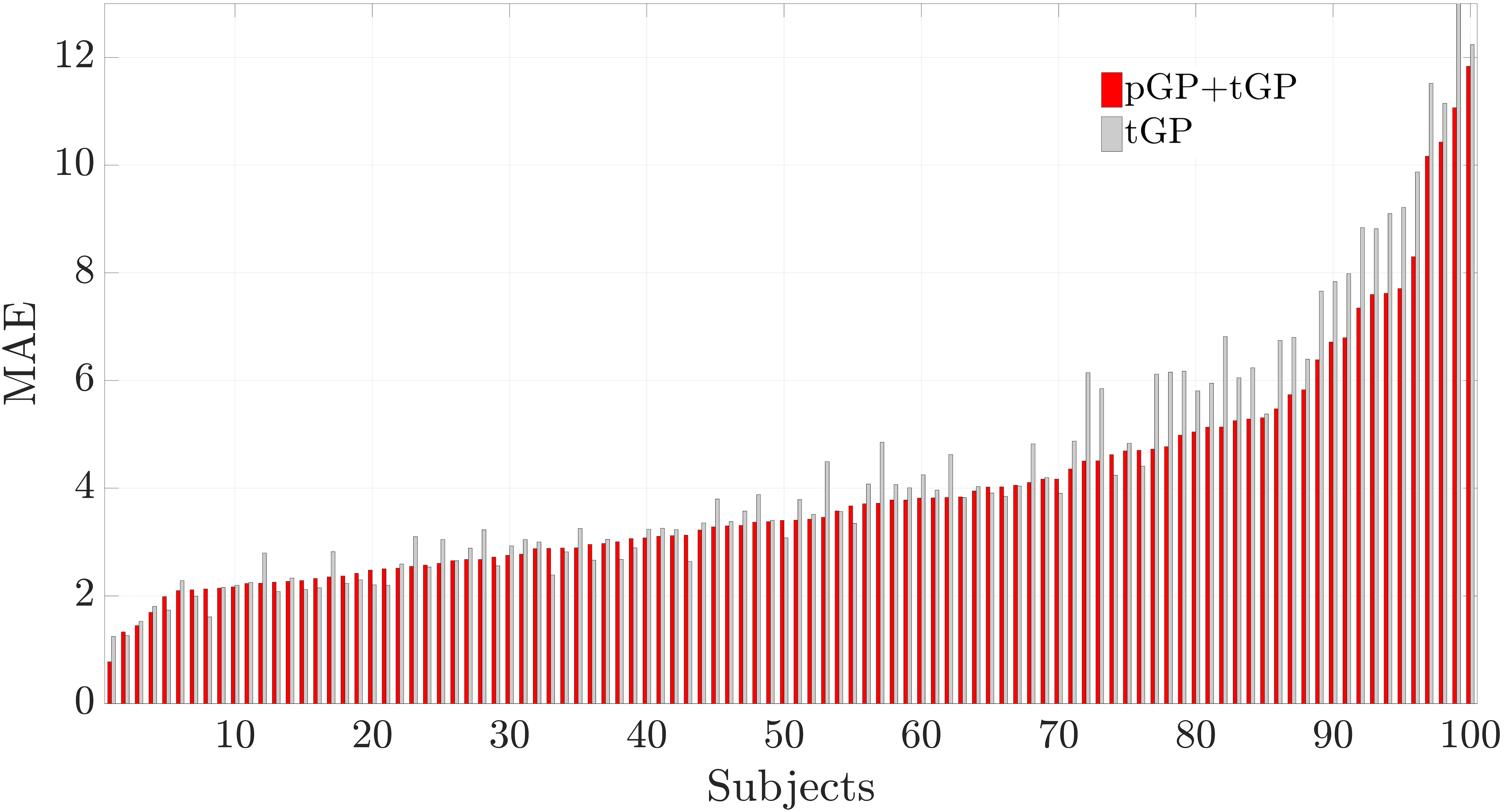}\label{fig:f6}}
  \caption{Each subject's average MAE over the forecasting window $w=\{6,12,18,24\}$ months. We contrast the joint pGP+tGP model against (a) sGP, (b) pGP, and (c) tGP, respectively. The subjects are sorted based on the increasing MAE of the pGP+tGP model.}
\end{figure*}

\newpage

\section{Conclusions}
\label{headings}
This work developed two variants of personalized GP models, both customized for the challenging task of simultaneously predicting four ADAS-Cog13 scores, at 6, 12, 18, and 24 months into the future. We tested these models on a cohort of 100 ADNI subjects, showing significant improvements over population-level GP models. Furthermore, we showed that by combining the personalized GP models (that use domain adaptation techniques) and subject-specific GP models, we achieve the best forecasting performance overall. This personalized forecasting framework enables prediction of changes in AD-related cognitive scores as early as possible, up to 24 months in the future, using a minimal set of subject data. This capability is of great importance to both clinicians and those at risk of AD since it is critical to early identification of at-risk subjects, construction of informative clinical trials, and timely detection of AD.

\section{Acknowledgments}
The work of O. Rudovic is funded by the European Union H2020, Marie Curie Action - Individual Fellowship no. 701236 (EngageMe).
\bibliography{egbib}

\begin{thebibliography}{10}
\providecommand{\url}[1]{#1}
\csname url@samestyle\endcsname
\providecommand{\newblock}{\relax}
\providecommand{\bibinfo}[2]{#2}
\providecommand{\BIBentrySTDinterwordspacing}{\spaceskip=0pt\relax}
\providecommand{\BIBentryALTinterwordstretchfactor}{4}
\providecommand{\BIBentryALTinterwordspacing}{\spaceskip=\fontdimen2\font plus
\BIBentryALTinterwordstretchfactor\fontdimen3\font minus
  \fontdimen4\font\relax}
\providecommand{\BIBforeignlanguage}[2]{{%
\expandafter\ifx\csname l@#1\endcsname\relax
\typeout{** WARNING: IEEEtran.bst: No hyphenation pattern has been}%
\typeout{** loaded for the language `#1'. Using the pattern for}%
\typeout{** the default language instead.}%
\else
\language=\csname l@#1\endcsname
\fi
#2}}
\providecommand{\BIBdecl}{\relax}
\BIBdecl

\bibitem{eleftheriadis2017}
S.~Eleftheriadis, O.~Rudovic, M.~P. Deisenroth, and M.~Pantic, ``Gaussian
  process domain experts for modeling of facial affect,'' \emph{IEEE
  Transactions on Image Processing}, vol.~26, no.~10, pp. 4697--4711, 2017.

\bibitem{mohs1997development}
R.~C. Mohs, D.~Knopman, R.~C. Petersen, S.~H. Ferris, C.~Ernesto, M.~Grundman,
  M.~Sano, L.~Bieliauskas, D.~Geldmacher, C.~Clark \emph{et~al.}, ``Development
  of cognitive instruments for use in clinical trials of antidementia drugs:
  additions to the alzheimer's disease assessment scale that broaden its
  scope.'' \emph{Alzheimer disease and associated disorders}, 1997.

\bibitem{skinner2012alzheimer}
J.~Skinner, J.~O. Carvalho, G.~G. Potter, A.~Thames, E.~Zelinski, P.~K. Crane,
  L.~E. Gibbons, A.~D.~N. Initiative \emph{et~al.}, ``The alzheimer’s disease
  assessment scale-cognitive-plus (adas-cog-plus): an expansion of the adas-cog
  to improve responsiveness in mci,'' \emph{Brain imaging and behavior},
  vol.~6, no.~4, pp. 489--501, 2012.

\bibitem{tadpole2017}
\BIBentryALTinterwordspacing
D.~Alexander, F.~Barkhof, F.~Nick, E.~Bron, and A.~Toga, ``The alzheimer's
  disease prediction of longitudinal evolution (tadpole) challenge,'' May 2017.
  [Online]. Available: \url{https://tadpole.grand-challenge.org/home/}
\BIBentrySTDinterwordspacing

\bibitem{cummings2006}
J.~L. Cummings, ``Challenges to demonstrating disease-modifying effects in
  alzheimer’s disease clinical trials,'' \emph{Alzheimer's \& Dementia},
  vol.~2, no.~4, pp. 263--271, 2006.

\bibitem{weiner2017}
M.~W. Weiner, D.~P. Veitch, P.~S. Aisen, L.~A. Beckett, N.~J. Cairns, R.~C.
  Green, D.~Harvey, C.~R. Jack, W.~Jagust, J.~C. Morris \emph{et~al.}, ``Recent
  publications from the alzheimer's disease neuroimaging initiative: Reviewing
  progress toward improved ad clinical trials,'' \emph{Alzheimer's \&
  Dementia}, 2017.

\bibitem{campos2015}
S.~Campos, L.~Pizarro, C.~Valle, K.~R. Gray, D.~Rueckert, and H.~Allende,
  ``Evaluating imputation techniques for missing data in adni: a patient
  classification study,'' in \emph{Iberoamerican Congress on Pattern
  Recognition}.\hskip 1em plus 0.5em minus 0.4em\relax Springer, 2015, pp.
  3--10.

\bibitem{rasmussen2006gaussian}
C.~Rasmussen and C.~Williams, \emph{Gaussian processes for machine
  learning}.\hskip 1em plus 0.5em minus 0.4em\relax MIT press {C}ambridge, MA,
  2006, vol.~1.

\bibitem{peterson2017personalized}
K.~Peterson, O.~Rudovic, R.~Guerrero, and R.~W. Picard, ``Personalized gaussian
  processes for future prediction of alzheimer's disease progression,''
  \emph{NIPS Workshop on Machine Learning for Healthcare (ML4HC)}, 2017.

\bibitem{liu2015bayesian}
B.~Liu and N.~Vasconcelos, ``Bayesian model adaptation for crowd counts,'' in
  \emph{2015 IEEE International Conference on Computer Vision (ICCV)}, Dec.
  2015, pp. 4175--4183.

\bibitem{schmidt-richberg16}
A.~Schmidt-Richberg, C.~Ledig, R.~Guerrero, H.~Molina-Abril, A.~Frangi, and
  D.~Rueckert, ``{Learning Biomarker Models for Progression Estimation of
  Alzheimer's Disease.}'' \emph{PloS one}, vol.~11, no.~4, Jan. 2016.

\bibitem{guerrero2016}
R.~Guerrero, A.~Schmidt-Richberg, C.~Ledig, T.~Tong, R.~Wolz, D.~Rueckert,
  A.~D. N.~I. (ADNI, and others), ``Instantiated mixed effects modeling of
  alzheimer's disease markers,'' \emph{NeuroImage}, vol. 142, pp. 113--125,
  2016.

\bibitem{gavidia2017}
G.~Gavidia-Bovadilla, S.~Kanaan-Izquierdo, M.~Matar{\'o}-Serrat,
  A.~Perera-Lluna, A.~D.~N. Initiative \emph{et~al.}, ``Early prediction of
  alzheimer’s disease using null longitudinal model-based classifiers,''
  \emph{PloS one}, vol.~12, no.~1, 2017.

\bibitem{schmidt_richberg15_a}
A.~Schmidt-Richberg, R.~Guerrero, C.~Ledig, H.~Molina-Abril, A.~F. Frangi,
  D.~Rueckert, and ADNI, ``{Multi-stage Biomarker Models for Progression
  Estimation in Alzheimer’s Disease},'' ser. Lecture Notes in Computer
  Science, vol. 9123, 2015, pp. 387--398.

\bibitem{long2017}
X.~Long, L.~Chen, C.~Jiang, L.~Zhang, A.~D.~N. Initiative \emph{et~al.},
  ``Prediction and classification of alzheimer disease based on quantification
  of mri deformation,'' \emph{PloS one}, vol.~12, no.~3, 2017.

\bibitem{minhas2017}
S.~Minhas, A.~Khanum, F.~Riaz, A.~Alvi, and S.~A. Khan, ``A nonparametric
  approach for mild cognitive impairment to ad conversion prediction: Results
  on longitudinal data,'' \emph{IEEE journal of biomedical and health
  informatics}, vol.~21, no.~5, pp. 1403--1410, 2017.

\bibitem{gaser2013}
C.~Gaser, K.~Franke, S.~Kl{\"o}ppel, N.~Koutsouleris, H.~Sauer, A.~D.~N.
  Initiative \emph{et~al.}, ``Brainage in mild cognitive impaired patients:
  predicting the conversion to alzheimer’s disease,'' \emph{PloS one},
  vol.~8, no.~6, 2013.

\bibitem{pereira2017}
T.~Pereira, A.~Mendon, F.~Ferreira, S.~Madeira, M.~Guerreiro \emph{et~al.},
  ``Towards a reliable prediction of conversion from mild cognitive impairment
  to alzheimer’s disease: stepwise learning using time windows,'' in
  \emph{Medical Informatics and Healthcare}, 2017, pp. 19--26.

\bibitem{grassi2018a}
M.~Grassi, G.~Perna, D.~Caldirola, K.~Schruers, R.~Duara, and D.~A.
  Loewenstein, ``A clinically-translatable machine learning algorithm for the
  prediction of alzheimer’s disease conversion in individuals with mild and
  premild cognitive impairment,'' \emph{Journal of Alzheimer's Disease}, no.
  Preprint, pp. 1--19.

\bibitem{candela2003propagation}
J.~Q. Candela, A.~Girard, J.~Larsen, and C.~E. Rasmussen, ``Propagation of
  uncertainty in bayesian kernel models-application to multiple-step ahead
  forecasting,'' in \emph{IEEE International Workshop on Neural Networks for
  Signal Processing}, vol.~2, 2003, pp. II--701.

\bibitem{deisenroth2015distributed}
M.~P. Deisenroth and J.~W. Ng, ``Distributed gaussian processes,'' \emph{arXiv
  preprint arXiv:1502.02843}, 2015.

\end{thebibliography}


\end{document}